\documentclass[pdflatex,sn-mathphys-num]{sn-jnl}

\usepackage{graphicx}%
\usepackage{multirow}%
\usepackage{amsmath,amssymb,amsfonts}%
\usepackage{amsthm}%
\usepackage{mathrsfs}%
\usepackage[title]{appendix}%
\usepackage{xcolor}%
\usepackage{textcomp}%
\usepackage{manyfoot}%
\usepackage{booktabs}%
\usepackage{algorithm}%
\usepackage{algorithmicx}%
\usepackage{algpseudocode}%
\usepackage{listings}%
\usepackage[font=small]{subcaption}
 \usepackage{bibunits}
\usepackage{geometry}

\theoremstyle{thmstyleone}%
\newtheorem{theorem}{Theorem}
\theoremstyle{thmstyletwo}%

\theoremstyle{thmstylethree}%
\begin{document}

\title[Discrete Differential Principle for Continuous Smooth Function Representation]{Discrete Differential Principle for Continuous Smooth Function Representation}


\author[1,2]{\fnm{Guoyou} \sur{Wang}}\email{gywang@mail.hust.edu.cn}
\equalcont{These authors contributed equally to this work.}

\author*[2]{\fnm{Yihua} \sur{Tan}}\email{yhtan@hust.edu.cn}
\equalcont{These authors contributed equally to this work.}

\author[2]{\fnm{Shiqi} \sur{Liu}}\email{shiqi.liu647@foxmail.com}
\equalcont{These authors contributed equally to this work.}

\affil[1]{\orgname{National Key Laboratory of Science \& Technology on Multi-Spectral Information Processing}}

\affil*[2]{\orgdiv{College of artificial intelligence and automation}, \orgname{Huazhong University of Science and Technology}, \orgaddress{\street{Luoyu Street}, \city{Wuhan}, \postcode{430070}, \state{Hubei}, \country{China}}}


\abstract{ 
Taylor's formula holds significant importance in function representation, such as solving differential difference equations, ordinary differential equations, partial differential equations, and further promotes applications in visual perception, complex control, fluid mechanics, weather forecasting and thermodynamics. However, the Taylor's formula suffers from the curse of dimensionality and error propagation during derivative computation in discrete situations. In this paper, we propose a new discrete differential operator to estimate derivatives and to represent continuous smooth function locally using the Vandermonde coefficient matrix derived from truncated Taylor series. Our method simultaneously computes all derivatives of orders less than the number of sample points, inherently mitigating error propagation. Utilizing equidistant uniform sampling, it achieves high-order accuracy while alleviating the curse of dimensionality. We mathematically establish rigorous error bounds for both derivative estimation and function representation, demonstrating tighter bounds for lower-order derivatives.  We extend our method to the two-dimensional case, enabling its use for multivariate derivative calculations. Experiments demonstrate the effectiveness and superiority of the proposed method compared to the finite forward difference method for derivative estimation and cubic spline and linear interpolation for function representation. Consequently, our technique offers broad applicability across domains  such as vision representation, feature extraction, fluid mechanics, and cross-media imaging.  }




\maketitle
\begin{bibunit}[unsrt]
Function representations have been widely used in numerous fields including visual perception\cite{Rosenblatt1957The}, fluid mechanics analysis\cite{lewalle1994wavelet}, complex system control\cite{komornik2005fourier}, weather forecasting\cite{el2014efficient} and  thermodynamic\cite{bale1975series}. Historically, functions were represented by solving equations for their roots. However, Galois proved that for equations of degree five or higher, no universal root-finding formula exists, making this approach impractical. In the early 19th century, Fourier series were adopted for function representation. Yet Fourier series capture only global behavior, not local features, and inefficiently represent non-periodic signals due to the requirement of an infinite frequency range. Alternative approaches use differential equations: domains are discretized into grids, and finite difference methods\cite{leveque1998finite} (or others\cite{bathe2007finite}) solve these equations. However, the governing differential equations are often unknown, grids must be infinitely dense (linking to the curse of dimensionality\cite{koppen2000curse}), and error propagation plagues high-order derivative approximations. Other methods—such as Lagrange interpolation, spline interpolation, least squares, and numerical integration—also exhibit inherent limitations.

Taylor's formula is a promising tool with many applications in function representation, such as solving differential-difference equations\cite{gulsu2006taylor}, ordinary differential equations\cite{corliss1982solving,barrio2011breaking}, and partial differential equations\cite{jentzen2011taylor}. However, Taylor's formula still suffers from the curse of dimensionality and error propagation in derivative calculations in discrete situations. To address the difficulty of computing derivatives, a discrete differential operator\cite{isshiki2011discrete} based on Taylor's formula offers an effective approach. However, the accuracy of this operator is questionable. To overcome the curse of dimensionality and error propagation, can we find a better differential operator that simultaneously estimates derivatives and represents functions with guaranteed error bounds?

We propose a novel discrete differential operator for continuous smooth function representation(see Figure~\ref{fig:Our method for derivative estimation and function   representation}) to mitigating the error propagation problem and the curse of dimensionality issue. It utilizes equidistant uniform sampling and has high-order accuracy, mitigating the curse of dimensionality issue. It is designed to simultaneously compute all derivatives of orders less than the number of sample points by using the Vandermonde coefficient matrix described by truncated Taylor series, mitigating the error propagation problem.

\begin{figure}[htb]
  \centering
  \includegraphics[width=\textwidth]{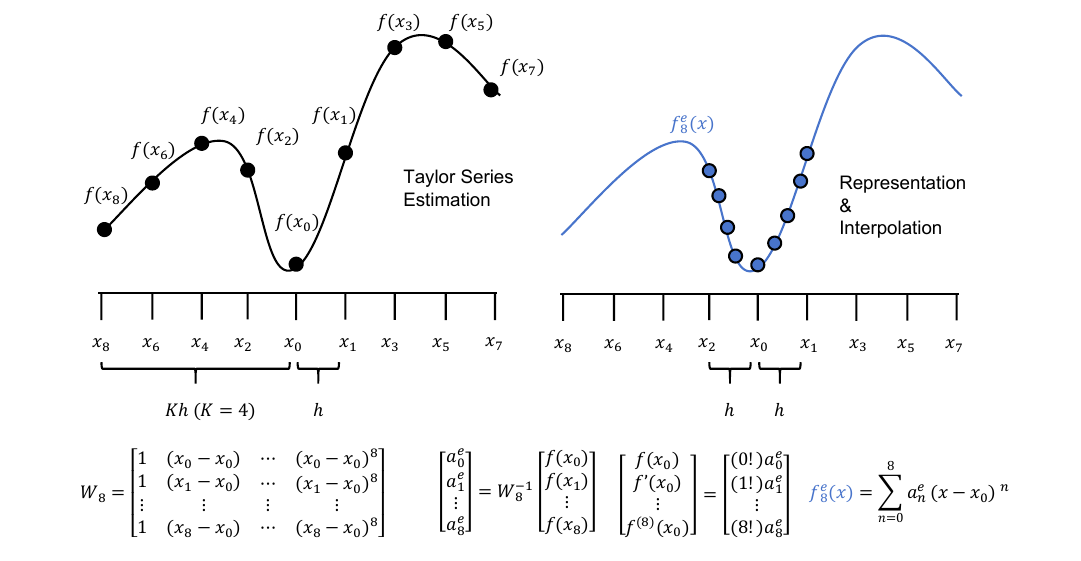}
  \caption{Our method for derivative estimation and function representation}\label{fig:Our method for derivative estimation and function representation}
\end{figure}

The following is an introduction to the discrete differential operator in the case of a single variable.
\subsection*{The differential operator for univariate functions}
To compute the derivatives located at $x_0$, suppose $x=x_0+h$, we could use Taylor's series to represent the function as follows:
\begin{equation}\label{eq:taylor's series}
  f(x)=\sum_{n=0}^{+\infty} a_n h^n
\end{equation}
where $a_n = \frac{f^{(n)}(x_0)}{n!}$.
We could truncate the equation to $N$, that is,
\begin{equation}\label{eq:taylor's series truncate}
  f(x)=(\sum_{n=0}^{N} a_n h^n) +R_N(h)
\end{equation}
where 
\begin{equation}\label{eq:R_N}
  R_N(x) = \frac{f^{(N+1)}(\zeta)}{(N+1)!}h^{N+1}
\end{equation} 
is the Lagrange form of the reminder and $\zeta$ is between $x_0$ and $x$.

Disregarding the the reminder, we could represent \begin{equation}\label{eqn:matrix representation}
                                                 f_N(x)=(1,h,\cdots,h^n)\begin{bmatrix}
                                                                       a_0 \\
                                                                       a_1 \\
                                                                       \vdots \\
                                                                       a_N 
                                                                     \end{bmatrix}.
                                               \end{equation}

In order to calculate $f^{(k)}(x_0)$, we could first calculate $a_k$. We could equidistantly uniform sample $x_1=(x_0+h_1),\cdots,x_{N+1}=(x_0+h_{N+1})$ from the neighborhood of $x_0$ and it yields
\begin{equation}\label{eqn:matrix form}
  \begin{bmatrix}
    f_N(x_1) \\
    f_N(x_2) \\
    \vdots \\
    f_N(x_{N+1}) 
  \end{bmatrix} = \begin{bmatrix}
                    1 & h_1 & \cdots & h_1^N \\
                    1 & h_2 & \cdots & h_2^N\\
                    \vdots & \vdots & \vdots & \vdots \\
                    1 & h_{N+1} & \cdots & h_{N+1}^N 
                  \end{bmatrix}\begin{bmatrix}
                                                                       a_0 \\
                                                                       a_1 \\
                                                                       \vdots \\
                                                                       a_N 
                                                                     \end{bmatrix}
\end{equation}
$W_N=\begin{bmatrix}
                    1 & h_1 & \cdots & h_1^N \\
                    1 & h_2 & \cdots & h_2^N\\
                    \vdots & \vdots & \vdots & \vdots \\
                    1 & h_{N+1} & \cdots & h_{N+1}^N 
                  \end{bmatrix}$ is Vandermonde matrix.
Then we have
\begin{equation}\label{eqn:estimate of ai}
  \begin{bmatrix}
                                                                       a^e_0 \\
                                                                       a^e_1 \\
                                                                       \vdots \\
                                                                       a^e_N 
                                                                     \end{bmatrix}=W_N^{-1}\begin{bmatrix}
                                                                                             f(x_1) \\
                                                                                             f(x_2) \\
                                                                                             \vdots \\
                                                                                             f(x_{N+1}) 
                                                                                           \end{bmatrix},
\end{equation}
where $a^e_i$ is the estimated coefficient for $a_i$ $i=0,\cdots,N$.
Then we have $ {f^{(n)}(x_0)}^e:=a_n^e n! $.
\subsection*{The differential operator for function representation}
After obtaining the estimated value of Taylor series, we could build up a function representation for $f(x)$ locally at neighborhood $[x_0-h,x_0+h]$, that is 
 \begin{equation}
                                                 f_N^e(x)=(1,x-x_0,\cdots,(x-x_0)^N)\begin{bmatrix}
                                                                       a^e_0 \\
                                                                       a^e_1 \\
                                                                       \vdots \\
                                                                       a^e_N 
                                                                     \end{bmatrix}
                                               \end{equation}
 where $N$ is the truncated number of Taylor series.
 
 The error, described in Theorem~\ref{thm:Errors of the function representation} in Appendix,  of this function representation is $O(h^{N+1} ((\frac{ K^{N+1}(K+1)^N }{N!})+ \frac{1}{(N+1)!}))$ where $h$ is the sample interval and $K$ is a multiple of the sampling area with respect to the sampling interval.  It implies that when $h$ is small, this representation can approximate the original function with very high-order precision.
 
 However, when the truncated number $N$ of Taylor series is high, we need to sample $N+1$ points to calculate the Vandermonde matrix and this may lead to extreme low value of the Vandermonde determinant(see Figure~\ref{figs: Comparison of number of sample points}(b) in Appendix), hindering the computation of inverse of the Vandermonde matrix. The experiments(see Figure~\ref{figs: Comparison of number of sample points}(a) in Appendix) also show that due to the increase of number of sample points, the errors from Vandermonde matrice increase and the errors of derivatives turn from decreasing to increasing.
 
 Therefore, we propose a multi-resolution encoding method to utilize a relatively small $N$ to achieve better representation capability. 
 Concretely, we utilize difference pyramid $D_1,D_2,\dots, D_m$ to represent the discrete sample points $F$ of the original signal $f(x)$. $F$ is the result of equally spaced sampling of $f(x)$.
 In order to introduce $D_1,D_2,\dots,D_m$, we first introduce filtered pyramid $G_1,G_2,\dots,G_m$. 
 
$G_1=2\times  down  sampling(F*\sigma)$,
  $G_2=2\times  down  sampling(G_1*\sigma)$,\dots,  $G_m=2\times  down  sampling(G_{m-1}*\sigma)$,
  where $\sigma$ is the filter kernel (mean or Gaussian kernel) and $*$ denotes the convolution operator.
  
$D_1 = F- 2\times upsampling(G_1)$, $D_2 = G_1-2\times upsampling(G_2)$, $\dots$, $D_{m-1} = G_{m-2}-2\times upsampling(G_{m-1})$, $D_m=G_m$ where we use the linear upsampling operator.
  
For each difference signal $D_i$, we select $x_0$ as the centered point and extract the $N+1$ points and their corresponding value ${x_1}_i=(x_0 + {h_1}_i),{x_2}_i=(x_0+{h_2}_i),\cdots,{x_{N+1}}_i=(x_0+{h_{N+1}}_i)$ from  the centered point's neighborhood. Then we construct the Vandermonde matrix,
${W_N}_{D_i}=\begin{bmatrix}
                    1 & {h_1}_i & \cdots & {h_1}_i^N \\
                    1 & {h_2}_i & \cdots & {h_2}_i^N\\
                    \vdots & \vdots & \vdots & \vdots \\
                    1 & {h_{N+1}}_i & \cdots & {h_{N+1}}_i^N 
                  \end{bmatrix}.$
we can obtain 
\begin{equation}\label{eqn:estimate of aid}
  \begin{bmatrix}
                                                                       {a^e_0}_{D_i} \\
                                                                       {a^e_1}_{D_i} \\
                                                                       \vdots \\
                                                                       {a^e_N}_{D_i} 
                                                                     \end{bmatrix}={W_N}_{D_i}^{-1}\begin{bmatrix}
                                                                                             D_i({x_1}_i) \\
                                                                                             D_i({x_2}_i) \\
                                                                                             \vdots \\
                                                                                             D_i({x_{N+1}}_i) 
                                                                                           \end{bmatrix},
\end{equation}
where $D_i(x)$ is the value of the sample signal $D_i$ at location $x$.

The final estimated Taylor series are $  \begin{bmatrix}
                                                                       {a^e_0} \\
                                                                       {a^e_1} \\
                                                                       \vdots \\
                                                                       {a^e_N} 
                                                                     \end{bmatrix}
                                                                     =\sum_{i=1}^{m}\begin{bmatrix}
                                                                       {a^e_0}_{D_i} \\
                                                                       {a^e_1}_{D_i} \\
                                                                       \vdots \\
                                                                       {a^e_N}_{D_i}  
                                                                     \end{bmatrix}$.
Empirically, we find that the multi-resolution encoding method can represent the signal better than the original differential operator method in the image domain and high-noise situations.

\subsection*{Contribution}
By calculating the inverse of Vandermonde matrix\cite{odeh1969art}, we show that in the univariate case, under proper conditions, the error of $i$-th order derivatives can be bounded by $O(\frac{ K^{2N+1-i} h^{N+1-i}}{(N-i)!})$ where $h$ is the sample interval, $N$ is the truncated index for Taylor series and $K$ is a multiple of the sampling area with respect to the sampling interval. We further show that the function representation error  is bounded locally by $O(h^{N+1} ((\frac{ K^{N+1}(K+1)^N }{N!})+ \frac{1}{(N+1)!}))$ under proper condition. We naturally generalize our method to the two-dimensional case so that it can be used for multivariate derivatives calculation. Our contributions are as follows:
 
\begin{itemize}
  \item We propose a novel discrete differential operator for representing continuous smooth functions. It maps any continuous smooth function to a low-dimensional manifold in an infinite-dimensional coefficient space, defined by the coefficients of its truncated Taylor series. The operator can calculate derivatives of different orders simultaneously. The estimated Taylor series coefficients can be used to represent the function locally, enabling new numerical methods for discrete signal analysis.
  \item We prove different error bounds for derivatives of different orders, and we also prove the function representation error bound in the one-dimensional case. 
  \item We generalize our method to compute derivatives of functions of two variables. This method can also locally represent two-variable functions.
  \item We conduct experiments demonstrating the effectiveness and superiority of our method in comparison to the finite forward difference method, cubic spline method, and linear interpolation method. 
\end{itemize}

\subsection*{Limitation}
Since we use Vandermonde matrix, we find that the interval for our methods can not be too small. Higher-order terms in the Vandermonde matrix and its inverse can cause numerical instability. Since the error bound is correlated with the number of sample points, the number of sample points should be in a proper interval. 

\subsection*{Conclusion}
We propose a novel discrete differential operator that can estimate derivatives of functions to various orders by using the Vandermonde coefficient matrix derived from truncated Taylor series. With the help of the discrete differential operator, we can also represent the different function locally. We mathematically prove the error bounds of derivatives and function representation. 
We then generalize our method to the two-dimensional case, enabling multivariate derivative calculations. Experiments validate the effectiveness and superiority of our method compared to the finite forward difference method for derivative estimation and cubic spline and linear interpolation for function representation.

\subsection*{Discussion}
Our method can be applied broadly across domains such as vision representation, feature extraction, fluid mechanics, and cross-media imaging. For instance, in image representation, we approximate the image function using coefficients from a multivariate truncated Taylor series. This approach simultaneously facilitates feature extraction by treating derivatives as interpretable features.
Using these extracted features, large neural networks can be made interpretable without the need for pre-training and it can alleviate the limitations of deep learning neural network models such as lack of interpretability, robustness and adaptability.

\putbib[sn-bibliography]
\end{bibunit}

\newpage
\section{Method}
\begin{bibunit}[unsrt]
We provide a proof sketch for the derivative error in the univariate case.
To estimate the error of our proposed method in the univariate case, we need to compute the inverse of the Vandermonde matrix. To calculate the inverse of the Vandermonde matrix
, for $y_1,\cdots,y_k$ we introduce elementary symmetric functions:
\begin{equation}\label{eqn:element symmetric function}
  e_m({y_1,\cdots,y_k})=\sum_{1\le j_1\le j2\le \cdots\le j_m\le k}y_{j_1}y_{j_2}\cdots y_{j_m}
\end{equation} for $m = 0 ,1,\cdots,k.$

Suppose $\{x_0,\cdots,x_N\}$ be a set of distinct values. Then
\cite{odeh1969art}
 \begin{equation}\label{eqn:inverse of Vandermonde matrix}
                                                                 (W_N)^{-1}_{(i+1)(j+1)}=\frac{(-1)^{N-i}e_{N-i}(\{x_0,\cdots,x_N\}\setminus\{x_j\})}{\prod_{m=0,m\neq j}^{N}(x_j-x_m)}
                                                               \end{equation} for $i,j=0,\cdots,N.$
Let
\begin{eqnarray}\label{eqn:estimation for residual}
  \begin{bmatrix}
                                                                       a^e_0 \\
                                                                       a^e_1 \\
                                                                       \vdots \\
                                                                       a^e_N 
                                                                     \end{bmatrix} &=& W_N^{-1}\begin{bmatrix}
                                                                                             f(x_0) \\
                                                                                             f(x_1) \\
                                                                                             \vdots \\
                                                                                             f(x_N) 
                                                                                           \end{bmatrix} = W_N^{-1}  \begin{bmatrix}
    f_N(x_0) +\frac{f^{(N+1)}(\zeta_0)}{(N+1)!}x_0^{N+1}\\
    f_N(x_1) +\frac{f^{(N+1)}(\zeta_1)}{(N+1)!}x_1^{N+1}\\
    \vdots \\
    f_N(x_N) +\frac{f^{(N+1)}(\zeta_N)}{(N+1)!}x_N^{N+1} 
  \end{bmatrix} \\
   &=& \begin{bmatrix}
                                                                       a_0 \\
                                                                       a_1 \\
                                                                       \vdots \\
                                                                       a_N 
                                                                     \end{bmatrix} +W_N^{-1}  \begin{bmatrix}
    \frac{f^{(N+1)}(\zeta_0)}{(N+1)!}x_0^{N+1}\\
    \frac{f^{(N+1)}(\zeta_1)}{(N+1)!}x_1^{N+1}\\
    \vdots \\
    \frac{f^{(N+1)}(\zeta_N)}{(N+1)!}x_N^{N+1} 
  \end{bmatrix}
\end{eqnarray}

Suppose $\vert x_i -x_j\vert\ge h$ for $i\neq j$ and $\vert x_i\vert \le Kh$ for $h>0$, $K>0$ and $i=0,\cdots, N$, we have 
\begin{equation}\label{eqn:inverse of Vandermonde matrix 1}
\vert (W_N)^{-1}_{(i+1)(j+1)}\vert \le \frac{C_{N}^{N-i}(Kh)^{N-i}}{h^N}= C_{N}^{N-i}(K)^{N-i}h^{-i}
                                                               \end{equation}

 Suppose $f^{(N+1)}(\zeta_k)$ is bounded by $M$ for $k=0,1,\cdots,N$.
 Therefore, it yields
 \begin{equation}\label{eqn:estimate for an}
  \vert a^e_i -a_i\vert \le \frac{M(Kh)^{N+1}}{(N+1)!} \times (N+1) \times C_{N}^{N-i}(K)^{N-i}h^{-i} = \frac{M C_N^{N-i}K^{2N+1-i} h^{N+1-i}}{N!}
 \end{equation}
 for $i=0,\cdots,N$.
 
 Therefore, it yields
 \begin{equation}\label{eqn:estimate for fi}
  \vert {f^{(i)}}^e(0) -{f^{(i)}(0)}\vert \le \frac{M K^{2N+1-i} h^{N+1-i}}{(N-i)!}.
 \end{equation}

 This leads to the following theorem:
 \begin{theorem}[Errors of the differential operators at zero\label{thm:Errors of the differential operators at zero}] Suppose $f$ is a smooth function and $\vert f^{(N+1)}(\zeta)\vert <M$ for $\vert \zeta\vert\le Kh$ where $K>0$ and $h>0$. We sample $N+1$ points $x_0,\cdots,x_N$ from the neighborhood of 0, satisfying $\vert x_i -x_j\vert\ge h$ $(\forall i\neq j)$ and $\vert x_i\vert \le Kh$ for $i=0,\cdots,N$. Let \begin{equation}\label{eq:taylor's series truncate}
  f(x)=(\sum_{n=0}^{N} a_n x^n) +R_N(x)
\end{equation}
where  $a_n = \frac{f^{(n)}(0)}{n!}$ and
\begin{equation}\label{eq:R_N}
  R_N(x) = \frac{f^{(N+1)}(\zeta)}{(N+1)!}x^{N+1}
\end{equation} 
is the Lagrange form of the reminder and $\zeta$ is between $0$ and $x$.  Let $W_N=\begin{bmatrix}
                    1 & x_0 & \cdots & x_0^N \\
                    1 & x_1 & \cdots & x_1^N \\
                    \vdots & \vdots & \vdots & \vdots \\
                    1 & x_N & \cdots & x_N^N 
                  \end{bmatrix}$ be Vandermonde matrix. Let \begin{equation}\label{eqn:estimation for residual simplified}
\begin{bmatrix}
                                                                       a^e_0 \\
                                                                       a^e_1 \\
                                                                       \vdots \\
                                                                       a^e_N 
                                                                     \end{bmatrix}=W_N^{-1}\begin{bmatrix}
                                                                                             f(x_0) \\
                                                                                             f(x_1) \\
                                                                                             \vdots \\
                                                                                             f(x_N) 
                                                                                           \end{bmatrix}.
\end{equation}Then we have  \begin{equation}\label{eqn:estimate for an simplified}
  \vert a^e_i -a_i\vert \le \frac{M C_N^{i}K^{2N+1-i} h^{N+1-i}}{N!}
 \end{equation} 
 for $i=0,\cdots,N$ and let ${{f^{(i)}}(0)}^e := i!a^e_i$, it yields \begin{equation}\label{eqn:estimate for fi}
  \vert {f^{(i)}(0)}^e -{f^{(i)}(0)}\vert \le \frac{M K^{2N+1-i} h^{N+1-i}}{(N-i)!}.
 \end{equation}
 \end{theorem}
 The proof of the theorem is in appendix~\ref{secA1}.
 According to the Theorem~\ref{thm:Errors of the differential operators}, our method can naturally compute all derivatives (at zero) of order less than or equal to $N$ at once. It also demonstrates that when the order of derivatives is small, the error bound is much more tight.  Notice that there are $K^{2N+1-i}$ in the error bound. Since we perform equidistant sampling, then $2K=N$, it yields $K=\frac{N}{2}$. The error bound turns to be $O(\frac{(\frac{N}{2})^{2N+1-i} h^{N+1-i}}{(N-i)!})$. We set $h=0.0625$ and we plot the relationship between $i$-th derivative's error bound and number of sample points. 
 
\begin{figure}[htb]
\centering
	\centering
\includegraphics[width=\textwidth]{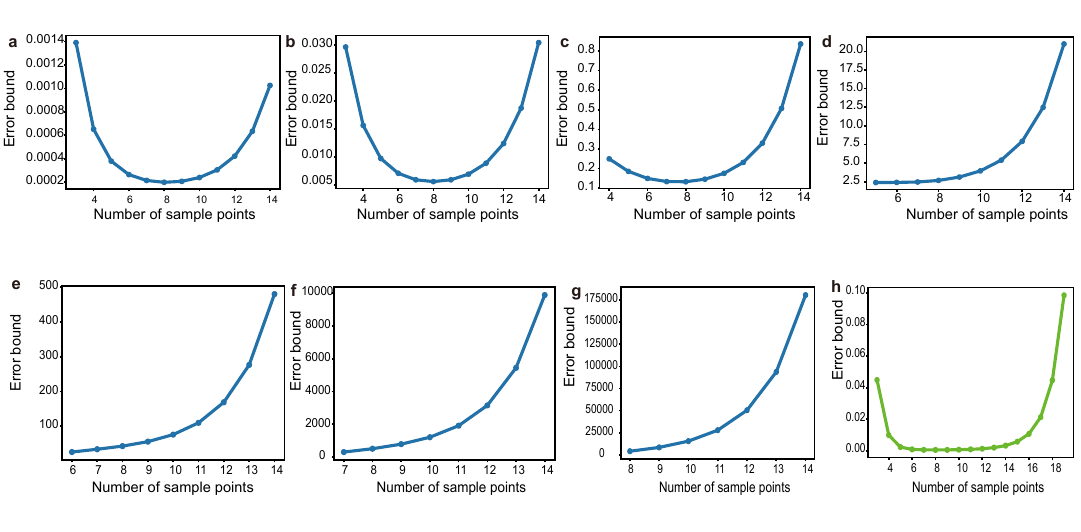}
\caption{The relationship between $i$-th derivative's ( and function representation) error bound and number of sample points.\label{fig: The relationship between $i$-th derivative's error bound and number of sample points}\textbf{a}. 1st derivative's error bound. \textbf{b}. 2nd derivative's error bound.  \textbf{c}. 3rd derivative's error bound.  \textbf{d}. 4th derivative's error bound. \textbf{e}. 5th derivative's error bound. \textbf{f}. 6th derivative's error bound. \textbf{g}. 7th derivative's error bound. \textbf{h}. Function representation error bound. }
\end{figure}

According to Figure~\ref{fig: The relationship between $i$-th derivative's error bound and number of sample points}, the errors of $1st,2nd,3rd,4th$ derivatives turn from decreasing to increasing with the increase in the number of sample points. The errors of $5th,6th,7th$ derivatives increase with the increase in the number of sample points. As a result, the best number of sample points in this parameter setting is around 7 to 9. 

When $h= 0.03125$, the optimal number of sample points is around 17 to 19. When $h\le 0.015625$, the errors of $1-7th$ derivatives  decrease with 3 to 35 sample points. We can conclude that for  a given small interval, the errors of derivatives decrease as the number of sample points increases  in a proper range.(However, due to the extremely low determinant of the Vandermonde matrix (see Figure~\ref{figs: Comparison of number of sample points}), which can lead to numerical instability, the number of sample points shouldn't be excessively large.)

We present an error theorem for differential operators at an arbitrary point (rather than at 0).
 \begin{theorem}[Errors of the differential operators\label{thm:Errors of the differential operators}] Suppose $f$ is a smooth function and $\vert f^{(N+1)}(\zeta)\vert <M$ for $\vert \zeta-x_0\vert\le Kh$ where $K>0$ and $h>0$. We sample $N+1$ points $x_1,\cdots,x_{N+1}$ from the neighborhood of 0, satisfying $\vert x_i -x_j\vert\ge h$ $(\forall i\neq j)$ and $\vert x_i-x_0\vert \le Kh$ for $i=1,\cdots,N+1$. Let \begin{equation}\label{eq:taylor's series truncate}
  f(x)=(\sum_{n=0}^{N} a_n (x-x_0)^n) +R_N(x)
\end{equation}
where  $a_n = \frac{f^{(n)}(x_0)}{n!}$ and
\begin{equation}\label{eq:R_N}
  R_N(x) = \frac{f^{(N+1)}(\zeta)}{(N+1)!}(x-x_0)^{N+1}
\end{equation} 
is the Lagrange form of the reminder and $\zeta$ is between $0$ and $x$. Let $h_i=x_i-x_0$ for $i=1,\cdots,N+1$. Let $W_N=\begin{bmatrix}
                    1 & h_1 & \cdots & h_1^N \\
                    1 & h_2 & \cdots & h_2^N \\
                    \vdots & \vdots & \vdots & \vdots \\
                    1 & h_{N+1} & \cdots & h_{N+1}^N 
                  \end{bmatrix}$ be Vandermonde matrix. Let \begin{equation}\label{eqn:estimation for residual simplified}
\begin{bmatrix}
                                                                       a^e_0 \\
                                                                       a^e_1 \\
                                                                       \vdots \\
                                                                       a^e_N 
                                                                     \end{bmatrix}=W_N^{-1}\begin{bmatrix}
                                                                                             f(x_0) \\
                                                                                             f(x_1) \\
                                                                                             \vdots \\
                                                                                             f(x_N) 
                                                                                           \end{bmatrix}.
\end{equation}Then we have  \begin{equation}\label{eqn:estimate for an simplified}
  \vert a^e_i -a_i\vert \le \frac{M C_N^{i}K^{2N+1-i} h^{N+1-i}}{N!}
 \end{equation} 
 for $i=0,\cdots,N$ and let ${{f^{(i)}}(x_0)}^e := i!a^e_i$, it yields \begin{equation}\label{eqn:estimate for fi}
  \vert {f^{(i)}(x_0)}^e -{f^{(i)}(x_0)}\vert \le \frac{M K^{2N+1-i} h^{N+1-i}}{(N-i)!}
 \end{equation}
 for $i=0,\cdots,N$.
 \end{theorem}
 
  The proof of the theorem is in appendix~\ref{secA1}.
 According to the Theorem~\ref{thm:Errors of the differential operators}, our method can naturally compute all derivatives of order less than or equal to $N$ at once. It also demonstrates that when the order of derivatives is small, the error bound is much more tight.
  Then we can present the theorem of errors of the function representation.

 \begin{theorem}[Errors of the function representation\label{thm:Errors of the function representation}] Suppose $f$ is a smooth function and $\vert f^{(N+1)}(\zeta)\vert <M$ for $\vert \zeta-x_0\vert\le Kh$ where $K>0$ and $h>0$. We sample $N+1$ points $x_1,\cdots,x_{N+1}$ from the neighborhood of 0, satisfying $\vert x_i -x_j\vert\ge h$ $(\forall i\neq j)$ and $\vert x_i-x_0\vert \le Kh$ for $i=1,\cdots,N+1$. Let \begin{equation}\label{eq:taylor's series truncate}
  f(x)=(\sum_{n=0}^{N} a_n (x-x_0)^n) +R_N(x)
\end{equation}
where  $a_n = \frac{f^{(n)}(x_0)}{n!}$ and
\begin{equation}\label{eq:R_N}
  R_N(x) = \frac{f^{(N+1)}(\zeta)}{(N+1)!}(x-x_0)^{N+1}
\end{equation} 
is the Lagrange form of the reminder and $\zeta$ is between $0$ and $x$. Let $h_i=x_i-x_0$ for $i=1,\cdots,N+1$. Let $W_N=\begin{bmatrix}
                    1 & h_1 & \cdots & h_1^N \\
                    1 & h_2 & \cdots & h_2^N \\
                    \vdots & \vdots & \vdots & \vdots \\
                    1 & h_{N+1} & \cdots & h_{N+1}^N 
                  \end{bmatrix}$ be Vandermonde matrix. Let \begin{equation}\label{eqn:estimation for residual simplified}
\begin{bmatrix}
                                                                       a^e_0 \\
                                                                       a^e_1 \\
                                                                       \vdots \\
                                                                       a^e_N 
                                                                     \end{bmatrix}=W_N^{-1}\begin{bmatrix}
                                                                                             f(x_0) \\
                                                                                             f(x_1) \\
                                                                                             \vdots \\
                                                                                             f(x_N) 
                                                                                           \end{bmatrix}.
\end{equation}
Let  \begin{equation}\label{eqn:matrix representation}
                                                 f_N^e(x)=(1,x-x_0,\cdots,(x-x_0)^N)\begin{bmatrix}
                                                                       a^e_0 \\
                                                                       a^e_1 \\
                                                                       \vdots \\
                                                                       a^e_N 
                                                                     \end{bmatrix}.
                                               \end{equation}
if $\vert x-x_0\vert \le h$, 
Then we have  \begin{equation}\label{eqn:estimate for function representation}
                  \vert f(x)-f_N^e(x) \vert \le  M h^{N+1} ((\frac{ K^{N+1}(K+1)^N }{N!})+ \frac{1}{(N+1)!}).
              \end{equation}
 \end{theorem}

   The proof of the theorem is in appendix~\ref{secA1}.
 According to the Theorem~\ref{thm:Errors of the function representation}, our method can represent the function locally with a very high-order precision. Notice that there are $((\frac{ K^{N+1}(K+1)^N }{N!})+ \frac{1}{(N+1)!})$ in the error bound. Since we perform equidistant sampling, then $2K=N$, it yields $K=\frac{N}{2}$. The error bound turns to be $O(h^{N+1}((\frac{ (\frac{N}{2})^{N+1}(\frac{N}{2}+1)^N }{N!})+ \frac{1}{(N+1)!}))$. We set $h=0.0625$ and we plot the relationship between  function representation's error bound and number of sample points. 
 

According to Figure~\ref{fig: The relationship between $i$-th derivative's error bound and number of sample points},  the error of function representation turns from decreasing to increasing with the increase in the number of sample points and the best number of sample points in this parameter setting is around 7 to 13.
 
When $h= 0.03125$, the optimal number of sample points is around 17. When $h\le 0.015625$, the function representation error  decreases with 3 to 35 sample points. We can conclude that for  a given small interval, the  function representation error decreases as the number of sample points increases in a proper range. (However, due to the extremely low determinant of the Vandermonde matrix (see Figure~\ref{figs: Comparison of number of sample points}), which can lead to numerical instability, the number of sample points shouldn't be excessively large.)

\putbib[sn-bibliography]
\end{bibunit}

\noindent\textbf{Author contribution} Guoyou Wang developed the discrete operator and multi-resolution encoding method for function representation. Shiqi Liu proved the error bounds of the function representation and derivative estimation. Yihua Tan supervised the work.

\noindent\textbf{Competing interests} The authors declare no competing interests.

\newpage
\setcounter{secnumdepth}{1}
\begin{appendices}
\section{The differential operator for multivariate functions}
For a multivariate function $f(x,y)$, suppose $x=x_0+h$ and $y=y_0+k$, we could use Taylor's series to represent the function as follows:
\begin{equation}\label{multivariate taylor series}
  f(x,y) = \sum_{i=0}^{N}\sum_{j=0}^{i} a_{i,j} h^j k^{i-j} + R_N
\end{equation}
where $a_{i,j}=C_i^j\frac{\partial^if(x_0,y_0)}{\partial x^j\partial y^{i-j}}$ and \begin{equation}\label{rn for 2}
                                                                                     R_N=(h\frac{\partial}{\partial x}+ k\frac{\partial}{\partial y})^{N+1}\frac{f(x+(\theta-1)h,y+(\theta-1)k)}{(N+1)!}
                                                                                   \end{equation} for some $\theta$ satisfying $0<\theta<1$. 
Suppose the derivatives of $f(x+(\theta-1)h,y+(\theta-1)k)$ are bounded by positive number $M$. Suppose $h=\rho \cos \alpha$ and $k=\rho \sin \alpha$, $\rho=\sqrt{h^2+k^2}$. It yields

\begin{equation}\label{eqn: residual bound}
  \vert R_N \vert\le \frac{M\rho^{N+1}}{(N+1)!}(\vert\cos \alpha \vert +\vert\sin\alpha\vert)^{N+1}\le \frac{2^{(N+1)/2}}{(N+1)!}M\rho^{N+1}.
\end{equation} 

Let \begin{equation}\label{eqn:f_N}
      f_N(x,y)= \sum_{i=0}^{N}\sum_{j=0}^{i} a_{i,j} h^j k^{i-j}.
    \end{equation}


\begin{figure*}[htbp]
\centering
\includegraphics[width=\linewidth]{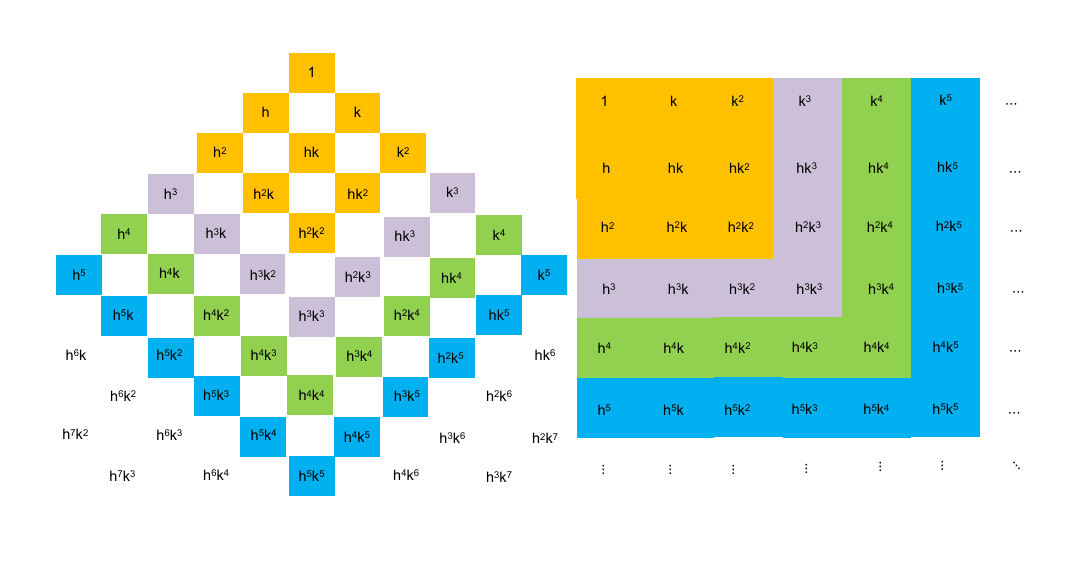}
\caption{The arrangement of the $h^jk^{i-j}$  term in the Taylor series.\label{fig:Zigzag }This arrangement minimizes the denominator of the first N items. }
\end{figure*}

As shown in Figure 1, we follow the ZigZag scanning order to form an m-dimensional column vector from the Taylor series terms $h^j k^{i-j}$.

Let $\Phi_m(h,k)=[\phi_0(h,k),\phi_1(h,k),\cdots,\phi_{m-1}(h,k)]^T$ be the Taylor series terms $h^jk^{i-j}$of the ZigZag scanning order. The corresponding coefficients $a_{i,j}$ form m-dimension column vector $G_m(x_0,y_0)=[g_0(x_0,y_0),g_1(x_0,y_0),\cdots,g_{m-1}(x_0,y_0)]^T$ where $g_l(x_0,y_0)=a_{i,j}=C_i^j\frac{\partial^if(x,y)}{\partial x^j\partial y^{i-j}}\vert_{(x_0,y_0)}$, $l=\frac{i^2+i}{2} +j$ and $m=\frac{n^2+3n}{2}$, $l=0,1,2,\cdots,m-1$.
It yields \begin{equation}\label{eqn:matrix representation simplified}
            f_N(x,y) = \Phi_m(h,k)^TG_m(x_0,y_0).
          \end{equation}
          
In order to calculate $G_m(x_0,y_0)$, we uniformly sample $m$ points from the neighborhood of $(x_0,y_0)$. Suppose the neighborhood of $(x_0,y_0)$ can be with grid size $N\times N$, $\Phi(h,k)$ takes $N\times N$ terms according to Figure~\ref{fig:Zigzag }. Then $m=N\times N$ and sample points are $(x_1,y_1)=(x_0,y_0)+(h_1,k_1),\cdots,(x_m,y_m)=(x_0,y_0)+(h_m,k_m)$, let $A_{m\times m}=\begin{bmatrix}
                         \Phi_m(h_1,k_1)^T \\
                         \Phi_m(h_2,k_2)^T \\
                         \vdots \\
                         \Phi_m(h_m,k_m)^T 
                       \end{bmatrix}.$
We can obtain
\begin{equation}\label{eqn:2 representation}
  F_m=\begin{bmatrix}
    f_N(x_1,y_1) \\
    f_N(x_2,y_2) \\
    \vdots \\
    f_N(x_m,y_m) 
  \end{bmatrix} = \begin{bmatrix}
                         \Phi_m(h_1,k_1)^T \\
                         \Phi_m(h_2,k_2)^T \\
                         \vdots \\
                         \Phi_m(h_m,k_m)^T 
                       \end{bmatrix}G_m(x_0,y_0)=A_{m\times m} G_m(x_0,y_0)
\end{equation}
 $A_{m\times m}$ is full rank $\iff$ for any  $F_m$, $G_m(x_0,y_0)$ has a unique solution. 

For example, if we take $m=9$, the range of the neighborhood $N(x_0,y_0)$ has a size of  $3\times 3$. $\Phi_9(h,k)=[1, h, k, h^2, hk, k^2,h^2k,hk^2,h^2k^2]$, it yields
\begin{equation}\label{eqn:example of m=9}
  F_9 = A_{9\times 9}G_9(x_0,y_0).
\end{equation}
We empirically find that $A_{9\times 9}$ has $9$ non-zero singular values, which means $A_{m\times m}$ is invertible. It yields
\begin{equation}\label{eqn:example of m=9 inverse}
  G_9(x_0,y_0) =A_{9\times 9}^{-1}G_9(x_0,y_0).
\end{equation} 
From Equation~\ref{eqn:example of m=9 inverse}, for any function $f(x,y)$ in the $3\times 3$ neighborhood of $(x_0,y_0)$, there is a unique solution for $G_9(x_0,y_0)$. In our experiments, as $N^2$ or $m$ increases, $A_{m\times m}$ always have $m$ non-zero singular values. As a result, $A_{m\times m}$ is invertible, and it yields
\begin{equation}\label{eqn: inverse}
  G_m(x_0,y_0) = A_{m\times m}^{-1}F_m.
\end{equation} 

Equation~\ref{eqn: inverse} demonstrates that we can recover the differential value by calculating $A_{m\times m}^{-1}F_m$. That is, $\frac{\partial^if(x,y)}{\partial x^j\partial y^{i-j}}\vert_{(x_0,y_0)}=\frac{g_l(x_0,y_0)}{C_i^j}$ where $l=\frac{i^2+i}{2} +j$.

Since $F_m$ is unknown, let \begin{equation}\label{eqn:2 representation r}
  F^r_m=\begin{bmatrix}
    f(x_1,y_1) \\
    f(x_2,y_2) \\
    \vdots \\
    f(x_m,y_m) 
  \end{bmatrix},\end{equation} it yields the estimation of $G^e_m(x_0,y_0)$ 
  \begin{equation}\label{eqn: estimate inverse}
  G_m^e(x_0,y_0) = A_{m\times m}^{-1}F_m^r.
\end{equation} 

Equation~\ref{eqn: estimate inverse} demonstrates that we can recover the estimated differential value by calculating $A_{m\times m}^{-1}F_m^r$. That is ${\frac{\partial^if(x,y)}{\partial x^j\partial y^{i-j}}\vert_{(x_0,y_0)}}^e=\frac{g_l^e(x_0,y_0)}{C_i^j}$ where $l=\frac{i^2+i}{2} +j$.

\section{Experiment}
To validate the theoretical analysis of our algorithm, we designed the experiments as follows:
\subsection{Experiments on exponential function}
\subsubsection{Derivative estimations}
To evaluate the performance of estimating the derivative values of different orders, we calculated 0-10th order derivative using 11 sample points with sample intervals of $0.5,0.25,0.125,0.0675,0.03375$. We chose $K=5$. For example, the 11 sample points with a 0.5 sample interval were $-2.5,-2, -1.5, -1, -0.5,  0,   0.5,  1,   1.5,  2,   2.5$ for our methods.

We select $f(x)=e^{2x}$ as the target function, and calculate its 0-10th order derivatives using both our proposed methods and the finite forward difference methods with same sample intervals. The results are as follows:

\begin{figure}
  \centering
  \caption{Derivative error comparison: Finite Difference vs. Our Method (applied to $e^{2x}$ and $(\sin x)\sin(10x)$)\label{tabs:Finite Difference vs. Our Method}. \textbf{a}. Error values for the finite forward difference applied to $e^{2x}$. \textbf{b}. Error values for our method applied to $e^{2x}$. \textbf{c}. Error values for the finite forward difference method applied to $(\sin x)\sin(10x)$. \textbf{d}. Error values for our method applied to $(\sin x)\sin(10x)$.}
  \includegraphics[width=\textwidth]{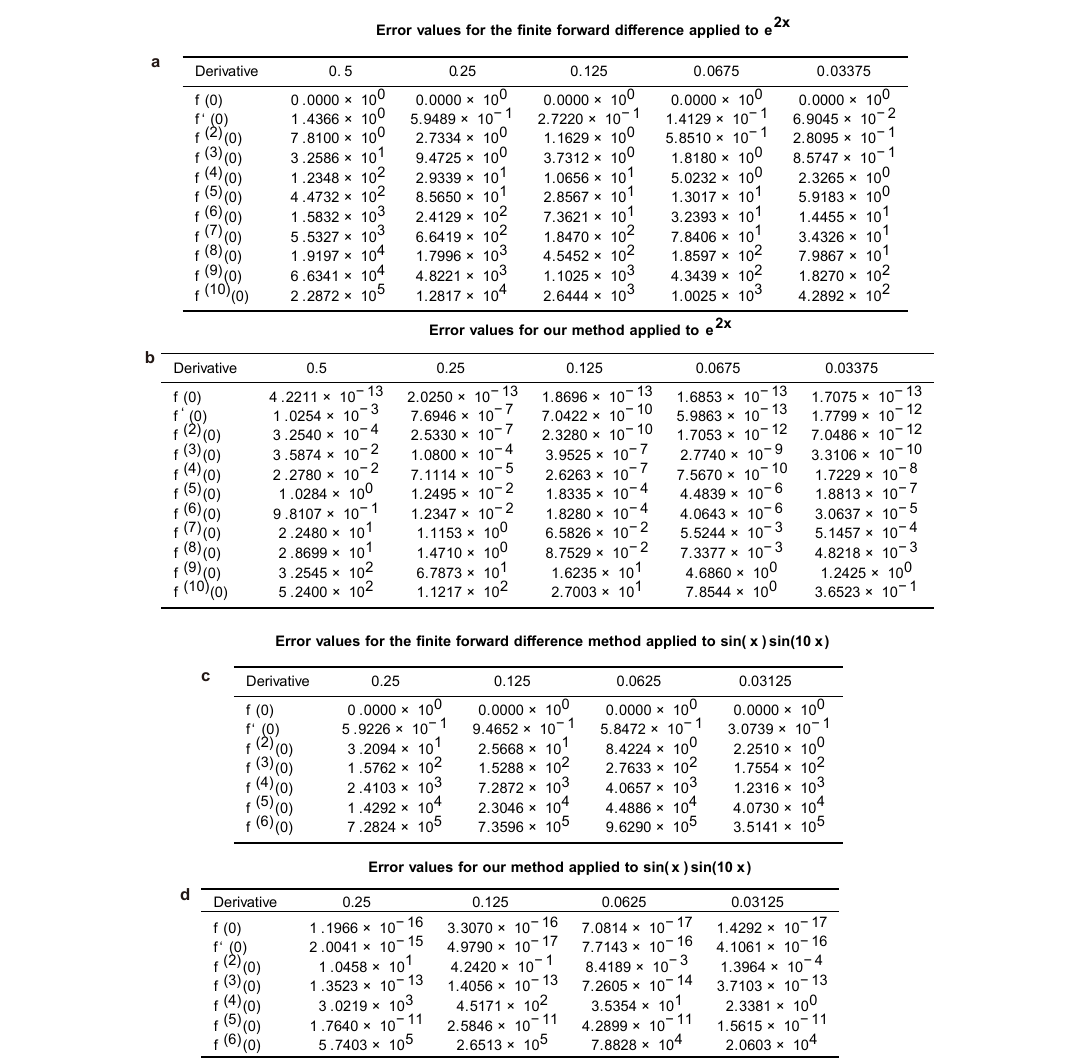}
\end{figure}


\begin{figure}[htbp]
\centering
\includegraphics[width=\textwidth]{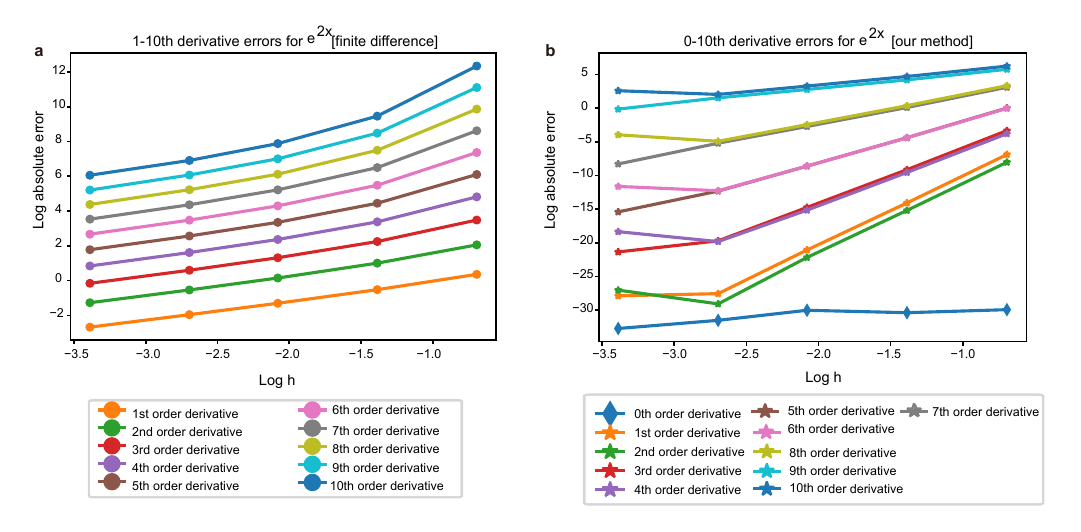}
\caption{Comparisons of errors. Since the $0$-th derivative of finite difference is always zero, it was not shown in figure. Acutally, our methods can also take $f(0)$ as the $0$-th derivative to achieve zero errors.\label{fig: Comparison of errors}}
\end{figure}

According to Figure~\ref{fig: Comparison of errors} and Table~\ref{tabs:Finite Difference vs. Our Method}, the error of our method is significantly smaller than that of the finite forward difference method. Besides, different order derivatives of the finite forward difference method may have similar 1 order error with respect to the interval $h$ while different order derivatives of our method have different order error with respect to the interval $h$. For example, $1$-st order derivative may have 8 order error with respect to the interval $h$ and $3$-rd order derivative may have 6 order error with respect to the interval $h$. The lower the order of the derivative, the higher the order of the error with respect to the interval $h$. 

\subsubsection{System error}
To demonstrate the influence of error from the system, we use different numbers of sample points with a sample interval of 0.125. The corresponding K values range from 1 to 10. For example, for our method, the 3 sample points with a 0.125 sample interval were -0.125, 0, and 0.125.

\begin{figure*}[htbp]
	\centering
\includegraphics[width=\textwidth]{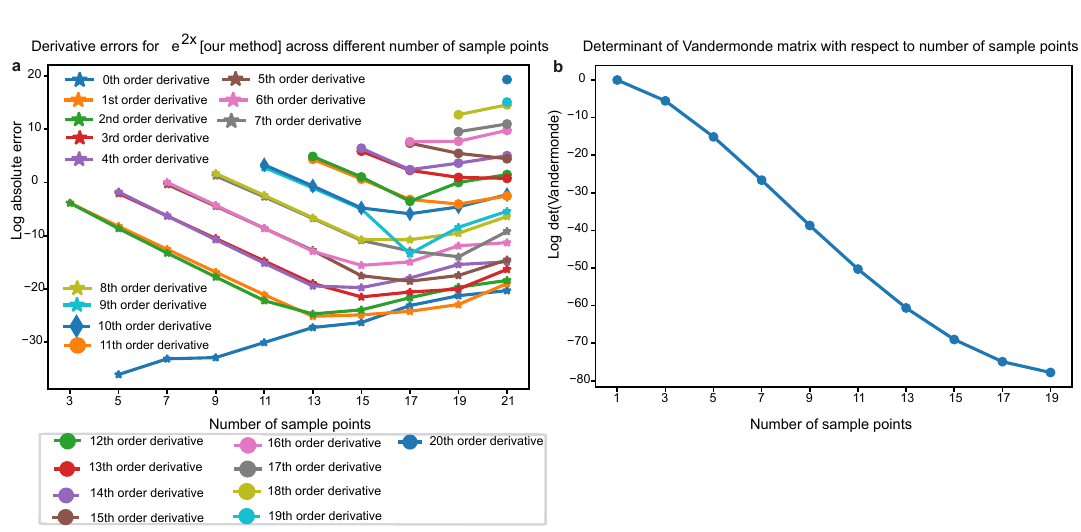}
\caption{Limitation of our method with respect to number of sample points. \textbf{a}. Derivative errors for $e^{2x}$ of our method across different number of sample points.\label{figs: Comparison of number of sample points} The $0th$ order derivative of three sample points is zero and is not shown in the figure. \textbf{b}. Determinants of the Vandermonde matrices.}
\end{figure*}

We plot the error of different derivatives with varying number of sample points in figure~\ref{figs: Comparison of number of sample points}. According to the figure, it is demonstrated that with the increase of sample points, the error of system becomes larger and the errors of derivatives turn from decreasing to increasing except for the $0-th$ derivative. This implies that we need to utilize a relative small number of sample points to construct the Vandermonde matrix to limit the error from the Vandermonde matrix and system's error bound.


\subsubsection{Function representation}
To evaluate the representation capability of our model, we first calculate discrete differential operator using 5 sample points with intervals 0.0625,0.03125,0.015625 at each sampled point(sample points start at -10, and total number of sample points is 300.) of the $e^{2x}$ , and then $4 \times$ interpolate function. The number of difference pyramid is 2 for the multi-resolution encoding method. We use cubic spline interpolation and linear interpolation as comparative methods.  We calculate the sum of the absolute error of each method's interpolation with the oracle function.

%
\begin{figure*}
  \centering
    \caption{Function representation error values for different methods applied to $e^{2x}$ and $(\sin x)\sin(10x)$. \textbf{a}. Function representation error values for different methods applied to $e^{2x}$. \textbf{b}. Function representation error values for different methods applied to $(\sin x) \sin(10x)$.}\label{tabs:Function representation error values for different methods.}
  \includegraphics[width=\textwidth]{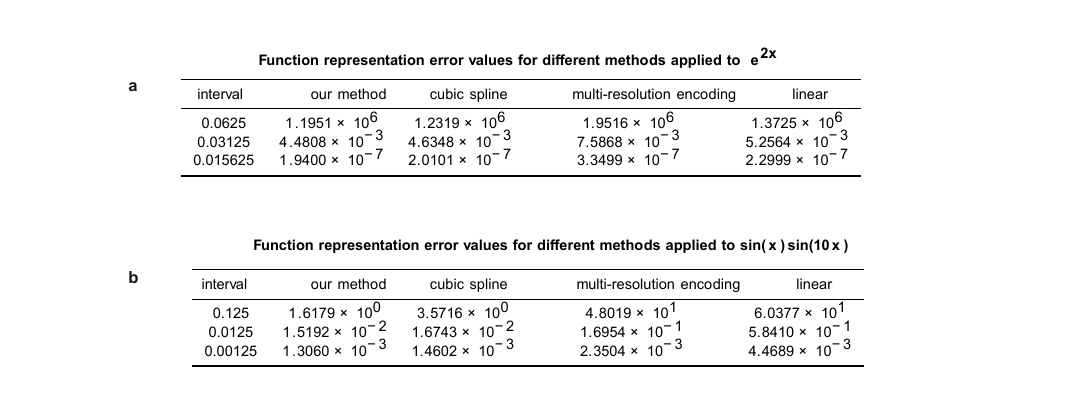}
\end{figure*}


\begin{figure}[htbp]
\centering
\includegraphics[width=\textwidth]{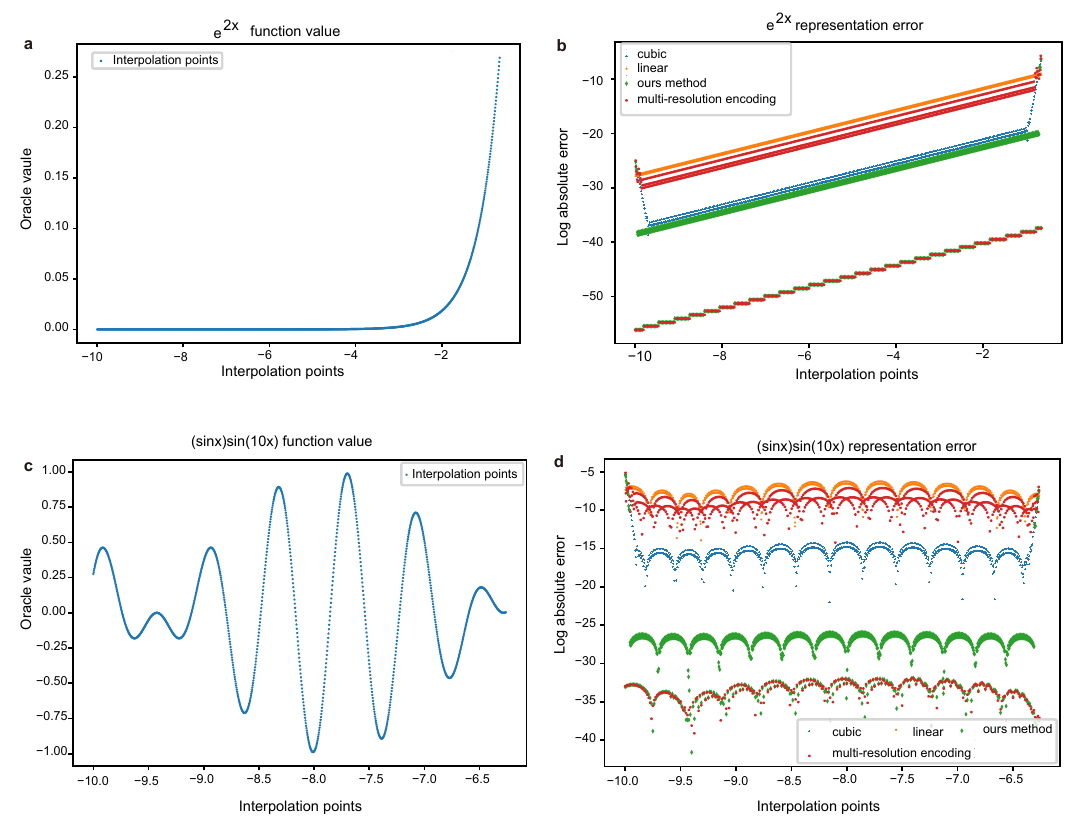}
\caption{Function value and representation error of $e^{2x}$ and $(\sin x)\sin(10x)$. \textbf{a}. Function value of $e^{2x}$. \textbf{b}. Comparisons of $(\sin x)\sin(10x)$ representation error (interval 0.0125). Since linear and cubic spline interpolation have zero errors in endpoints, their errors at endpoints are not shown in figure. Acutally, our methods can also take  endpoints values to achieve zero errors. \textbf{c}. Function value of $(\sin x)\sin (10x)$. \textbf{d}. Comparisons of $e^{2x}$ representation error (interval 0.03125). Since linear and cubic spline interpolation have zero errors in endpoints, their errors at endpoints are not shown in figure. \label{fig: Comparisons of function representation error}}
\end{figure}

According to Table~\ref{tabs:Function representation error values for different methods.} and Figure~\ref{fig: Comparisons of function representation error}, our vanilla method achieve best performance across different sample interval and cubic spline interpolation method is the second-best method. Our multi-resolution encoding method doesn't perform well and the possible reason is that the $e^{2x}$ signal are relative simple and the difference pyramid is not suited to represent the key information of it.

\subsection{Experiments on double sideband suppressed carrier modulated signal}
\subsubsection{Derivative estimations}
To evaluate the performance of estimating the derivative values of different orders, we calculated 0-6th order derivative with 7 sample points with sample intervals being $0.25,0.125,0.0625,0.03125$. We chose $K=3$. For example, the 7 sample points with a 0.5 sample interval were $-1.5, -1.0,  -0.5,  0.0,   0.5,  1.0,   1.5$ for our method.

We select $f(x)=\sin(x)\sin(10x)$ as the target function, and calculate its 0-6th order derivatives through our method as well as the finite forward difference method with same sample intervals. The results are as follows:

According to Table~\ref{tabs:Finite Difference vs. Our Method}, the error of our method is generally significantly smaller than that of the finite forward difference except for $f^{(4)}(0)$ with an interval of 0.25.  We notice that the $0$-th order derivatives of the finite forward difference method have zero errors because  no calculation is needed for $0$-th order derivatives. Similarly, if we set $f(0)$ to be the $0$-th order derivative in our method, then we would also achieve zero error.

\subsubsection{Function representation}
To evaluate the representation capability of our model, we first calculate discrete differential operator using 9 sample points with intervals 0.125,0.0125,0.00125 at each sampled point(sample points start at -10, and total number of sample points is 300.) of the $\sin(x)\sin(10x)$ , and then $4 \times$ interpolate function. The number of difference pyramid is 2 for the multi-resolution encoding method. We use cubic spline interpolation and linear interpolation as comparative methods.  We calculate the sum of the absolute error of each method's interpolation with the oracle function.

%


According to Table~\ref{tabs:Function representation error values for different methods.} and Figure~\ref{fig: Comparisons of function representation error}, our vanilla method achieve best performance across different sample interval and cubic spline interpolation method is the second-best method. Our multi-resolution encoding method doesn't perform well and the possible reason is that the 1 dimension signal are relative simple and the difference pyramid is not suited to represent the key information of it. The linear method is the most simple and worst method in the comparison.  

\subsection{The determinant of the Vandermonde matrix}

In this experiment, we calculated the determinant of the Vandermonde matrix to demonstrate that as the number of sample points increases, the determinant becomes very small and numerically unstable for inverse calculation. The sample interval is 0.125, and the number of sample points ranges from 1 to 19 in increments of 2. The results are in Figure~\ref{figs: Comparison of number of sample points}.

\section{Proof}\label{secA1}
\noindent\textbf{Theorem~\ref{thm:Errors of the differential operators at zero}}[Errors of the differential operators] Suppose $f$ is a smooth function and $\vert f^{(N+1)}(\zeta)\vert <M$ for $\vert \zeta\vert\le Kh$ where $K>0$ and $h>0$. We sample $N+1$ points $x_0,\cdots,x_N$ from the neighborhood of 0, satisfying $\vert x_i -x_j\vert\ge h$ $(\forall i\neq j)$ and $\vert x_i\vert \le Kh$ for $i=0,\cdots,N$. Let \begin{equation}\label{eq:taylor's series truncate}
  f(x)=(\sum_{n=0}^{N} a_n x^n) +R_N(x)
\end{equation}
where  $a_n = \frac{f^{(n)}(0)}{n!}$ and
\begin{equation}\label{eq:R_N}
  R_N(x) = \frac{f^{(N+1)}(\zeta)}{(N+1)!}x^{N+1}
\end{equation} 
is the Lagrange form of the reminder and $\zeta$ is between $0$ and $x$.  Let $W_N=\begin{bmatrix}
                    1 & x_0 & \cdots & x_0^N \\
                    1 & x_1 & \cdots & x_1^N \\
                    \vdots & \vdots & \vdots & \vdots \\
                    1 & x_N & \cdots & x_N^N 
                  \end{bmatrix}$ be the Vandermonde matrix. Let \begin{equation}\label{eqn:estimation for residual simplified}
\begin{bmatrix}
                                                                       a^e_0 \\
                                                                       a^e_1 \\
                                                                       \vdots \\
                                                                       a^e_N 
                                                                     \end{bmatrix}=W_N^{-1}\begin{bmatrix}
                                                                                             f(x_0) \\
                                                                                             f(x_1) \\
                                                                                             \vdots \\
                                                                                             f(x_N) 
                                                                                           \end{bmatrix}.
\end{equation}Then we have  \begin{equation}\label{eqn:estimate for an simplified}
  \vert a^e_i -a_i\vert \le \frac{M C_N^{i}K^{2N+1-i} h^{N+1-i}}{N!}
 \end{equation} 
 for $i=0,\cdots,N$ and let ${f^{(i)}(0)}^e := i!a^e_i$, it yields \begin{equation}\label{eqn:estimate for fi}
  \vert {f^{(i)}(0)}^e -{f^{(i)}(0)}\vert \le \frac{M K^{2N+1-i} h^{N+1-i}}{(N-i)!}.
 \end{equation}

 \begin{proof}\begin{equation}\label{eqn:estimation for residual}
\begin{bmatrix}
                                                                       a^e_0 \\
                                                                       a^e_1 \\
                                                                       \vdots \\
                                                                       a^e_N 
                                                                     \end{bmatrix} = W_N^{-1}  \begin{bmatrix}
    f_N(x_0) +\frac{f^{(N+1)}(\zeta_0)}{(N+1)!}x_0^{N+1}\\
    f_N(x_1) +\frac{f^{(N+1)}(\zeta_1)}{(N+1)!}x_1^{N+1}\\
    \vdots \\
    f_N(x_N) +\frac{f^{(N+1)}(\zeta_N)}{(N+1)!}x_N^{N+1} 
  \end{bmatrix}=\begin{bmatrix}
                                                                       a_0 \\
                                                                       a_1 \\
                                                                       \vdots \\
                                                                       a_N 
                                                                     \end{bmatrix} +W_N^{-1}  \begin{bmatrix}
    \frac{f^{(N+1)}(\zeta_0)}{(N+1)!}x_0^{N+1}\\
    \frac{f^{(N+1)}(\zeta_1)}{(N+1)!}x_1^{N+1}\\
    \vdots \\
    \frac{f^{(N+1)}(\zeta_N)}{(N+1)!}x_N^{N+1} 
  \end{bmatrix}.
\end{equation}
Therefore, \begin{equation}\label{eqn:ai estimation complex}
             \vert a^e_i -a_i\vert \le \vert {W_N^{-1}}_{i+1} \begin{bmatrix}
    \frac{f^{(N+1)}(\zeta_0)}{(N+1)!}x_0^{N+1}\\
    \frac{f^{(N+1)}(\zeta_1)}{(N+1)!}x_1^{N+1}\\
    \vdots \\
    \frac{f^{(N+1)}(\zeta_N)}{(N+1)!}x_N^{N+1} 
  \end{bmatrix}\vert \le \sum_{j=0}^{N} \vert (W_N)^{-1}_{(i+1)(j+1)}\vert \vert \frac{f^{(N+1)}(\zeta_j)}{(N+1)!}x_j^{N+1}\vert .
           \end{equation} where ${W_N^{-1}}_{i+1} $ is the $i+1$-th row of the matrix $W_N^{-1}$.
           
It yields \begin{eqnarray}\label{eqn:ai estimation complex 1}
              \vert a^e_i -a_i\vert &\le&  \sum_{j=0}^{N} \vert (W_N)^{-1}_{(i+1)(j+1)}\vert \vert \frac{f^{(N+1)}(\zeta_j)}{(N+1)!}x_j^{N+1}\vert  \\
              &\le& (N+1)  C_{N}^{N-i}(K)^{N-i}h^{-i} \times \frac{M(Kh)^{N+1}}{(N+1)!}  \\
              &=& \frac{M C_N^{i}K^{2N+1-i} h^{N+1-i}}{N!} 
           \end{eqnarray}
for $i=0,\cdots, N$. 

\begin{equation}\label{eqn:estimate for fi}
  \vert {{f^{(i)}}(0)}^e -{f^{(i)}(0)}\vert = i! \vert a^e_i -a_i\vert \le \frac{M C_N^{i}K^{2N+1-i} h^{N+1-i}i!}{N!} = \frac{M K^{2N+1-i} h^{N+1-i}}{(N-i)!}.
 \end{equation}               
 \end{proof}

\noindent\textbf{Theorem~\ref{thm:Errors of the differential operators}}[Errors of the differential operators] Suppose $f$ is a smooth function and $\vert f^{(N+1)}(\zeta)\vert <M$ for $\vert \zeta-x_0\vert\le Kh$ where $K>0$ and $h>0$. We sample $N+1$ points $x_1,\cdots,x_{N+1}$ from the neighborhood of 0, satisfying $\vert x_i -x_j\vert\ge h$ $(\forall i\neq j)$ and $\vert x_i-x_0\vert \le Kh$ for $i=1,\cdots,N+1$. Let \begin{equation}\label{eq:taylor's series truncate}
  f(x)=(\sum_{n=0}^{N} a_n (x-x_0)^n) +R_N(x)
\end{equation}
where  $a_n = \frac{f^{(n)}(x_0)}{n!}$ and
\begin{equation}\label{eq:R_N}
  R_N(x) = \frac{f^{(N+1)}(\zeta)}{(N+1)!}(x-x_0)^{N+1}
\end{equation} 
is the Lagrange form of the reminder and $\zeta$ is between $0$ and $x$. Let $h_i=x_i-x_0$ for $i=1,\cdots,N+1$. Let $W_N=\begin{bmatrix}
                    1 & h_1 & \cdots & h_1^N \\
                    1 & h_2 & \cdots & h_2^N \\
                    \vdots & \vdots & \vdots & \vdots \\
                    1 & h_{N+1} & \cdots & h_{N+1}^N 
                  \end{bmatrix}$ be Vandermonde matrix. Let \begin{equation}\label{eqn:estimation for residual simplified}
\begin{bmatrix}
                                                                       a^e_0 \\
                                                                       a^e_1 \\
                                                                       \vdots \\
                                                                       a^e_N 
                                                                     \end{bmatrix}=W_N^{-1}\begin{bmatrix}
                                                                                             f(x_0) \\
                                                                                             f(x_1) \\
                                                                                             \vdots \\
                                                                                             f(x_N) 
                                                                                           \end{bmatrix}.
\end{equation}Then we have  \begin{equation}\label{eqn:estimate for an simplified}
  \vert a^e_i -a_i\vert \le \frac{M C_N^{i}K^{2N+1-i} h^{N+1-i}}{N!}
 \end{equation} 
 for $i=0,\cdots,N$ and let ${{f^{(i)}}(x_0)}^e := i!a^e_i$, it yields \begin{equation}\label{eqn:estimate for fi}
  \vert {f^{(i)}(x_0)}^e -{f^{(i)}(x_0)}\vert \le \frac{M K^{2N+1-i} h^{N+1-i}}{(N-i)!}
 \end{equation}
for $i=0,\cdots,N.$

 \begin{proof}\begin{equation}\label{eqn:estimation for residual}
\begin{bmatrix}
                                                                       a^e_0 \\
                                                                       a^e_1 \\
                                                                       \vdots \\
                                                                       a^e_N 
                                                                     \end{bmatrix} = W_N^{-1}  \begin{bmatrix}
    f_N(x_1) +\frac{f^{(N+1)}(\zeta_0)}{(N+1)!}h_1^{N+1}\\
    f_N(x_2) +\frac{f^{(N+1)}(\zeta_1)}{(N+1)!}h_2^{N+1}\\
    \vdots \\
    f_N(x_{N+1}) +\frac{f^{(N+1)}(\zeta_N)}{(N+1)!}h_{N+1}^{N+1} 
  \end{bmatrix}=\begin{bmatrix}
                                                                       a_0 \\
                                                                       a_1 \\
                                                                       \vdots \\
                                                                       a_N 
                                                                     \end{bmatrix} +W_N^{-1}  \begin{bmatrix}
    \frac{f^{(N+1)}(\zeta_0)}{(N+1)!}h_1^{N+1}\\
    \frac{f^{(N+1)}(\zeta_1)}{(N+1)!}h_2^{N+1}\\
    \vdots \\
    \frac{f^{(N+1)}(\zeta_N)}{(N+1)!}h_{N+1}^{N+1} 
  \end{bmatrix}
\end{equation}
where \begin{equation}\label{eqn:matrix representation}
                                                 f_N(x)=(1,x-x_0,\cdots,(x-x_0)^n)\begin{bmatrix}
                                                                       a_0 \\
                                                                       a_1 \\
                                                                       \vdots \\
                                                                       a_N 
                                                                     \end{bmatrix}.
                                               \end{equation}
Therefore, \begin{equation}\label{eqn:ai estimation complex}
             \vert a^e_i -a_i\vert \le \vert {W_N^{-1}}_{i+1} \begin{bmatrix}
    \frac{f^{(N+1)}(\zeta_0)}{(N+1)!}h_1^{N+1}\\
    \frac{f^{(N+1)}(\zeta_1)}{(N+1)!}h_2^{N+1}\\
    \vdots \\
    \frac{f^{(N+1)}(\zeta_N)}{(N+1)!}h_{N+1}^{N+1} 
  \end{bmatrix}\vert \le \sum_{j=0}^{N} \vert (W_N)^{-1}_{(i+1)(j+1)}\vert \vert \frac{f^{(N+1)}(\zeta_j)}{(N+1)!}h_{j+1}^{N+1}\vert .
           \end{equation} where ${W_N^{-1}}_{i+1} $ is the $i+1$-th row of the matrix $W_N^{-1}$.

To calculate the inverse of the Vandermonde matrix\cite{odeh1969art}, for $y_1,\cdots,y_k$ we introduce elementary symmetric functions:
\begin{equation}\label{eqn:element symmetric function}
  e_m({y_1,\cdots,y_k})=\sum_{1\le j_1\le j2\le \cdots\le j_m\le k}y_{j_1}y_{j_2}\cdots y_{j_m}
\end{equation} for $m = 0 ,1,\cdots,k.$

Suppose $\{h_1,\cdots,h_{N+1}\}$ be a set of distinct values. Then\cite{odeh1969art} \begin{equation}\label{eqn:inverse of Vandermonde matrix}
                                                                 (W_N)^{-1}_{(i+1)(j+1)}=\frac{(-1)^{N-i}e_{N-i}(\{h_1,\cdots,h_{N+1}\}\setminus\{h_{j+1}\})}{\prod_{m=0,m\neq j}^{N}(h_{j+1}-h_{m+1})}
                                                               \end{equation} for $i,j=0,\cdots,N.$

\begin{equation}\label{eqn:inverse of Vandermonde matrix 1}
\vert (W_N)^{-1}_{(i+1)(j+1)}\vert \le \frac{C_{N}^{N-i}(Kh)^{N-i}}{h^N}= C_{N}^{N-i}(K)^{N-i}h^{-i}
                                                               \end{equation}

It yields \begin{eqnarray}\label{eqn:ai estimation complex 1}
              \vert a^e_i -a_i\vert &\le&  \sum_{j=0}^{N} \vert (W_N)^{-1}_{(i+1)(j+1)}\vert \vert \frac{f^{(N+1)}(\zeta_j)}{(N+1)!}h_{j+1}^{N+1}\vert  \\
              &\le& (N+1)  C_{N}^{N-i}(K)^{N-i}h^{-i} \times \frac{M(Kh)^{N+1}}{(N+1)!}  \\
              &=& \frac{M C_N^{i}K^{2N+1-i} h^{N+1-i}}{N!} 
           \end{eqnarray}
for $i=0,\cdots, N$. 

\begin{equation}\label{eqn:estimate for fi}
  \vert {{f^{(i)}}(x_0)}^e -{f^{(i)}(x_0)}\vert = i! \vert a^e_i -a_i\vert \le \frac{M C_N^{i}K^{2N+1-i} h^{N+1-i}i!}{N!} = \frac{M K^{2N+1-i} h^{N+1-i}}{(N-i)!}.
 \end{equation}               
 \end{proof}

\noindent\textbf{Theorem}~\ref{thm:Errors of the function representation}[Errors of the function representation] Suppose $f$ is a smooth function and $\vert f^{(N+1)}(\zeta)\vert <M$ for $\vert \zeta-x_0\vert\le Kh$ where $K>0$ and $h>0$. We sample $N+1$ points $x_1,\cdots,x_{N+1}$ from the neighborhood of 0, satisfying $\vert x_i -x_j\vert\ge h$ $(\forall i\neq j)$ and $\vert x_i-x_0\vert \le Kh$ for $i=1,\cdots,N+1$. Let \begin{equation}\label{eq:taylor's series truncate}
  f(x)=(\sum_{n=0}^{N} a_n (x-x_0)^n) +R_N(x)
\end{equation}
where  $a_n = \frac{f^{(n)}(x_0)}{n!}$ and
\begin{equation}\label{eq:R_N}
  R_N(x) = \frac{f^{(N+1)}(\zeta)}{(N+1)!}(x-x_0)^{N+1}
\end{equation} 
is the Lagrange form of the reminder and $\zeta$ is between $0$ and $x$. Let $h_i=x_i-x_0$ for $i=1,\cdots,N+1$. Let $W_N=\begin{bmatrix}
                    1 & h_1 & \cdots & h_1^N \\
                    1 & h_2 & \cdots & h_2^N \\
                    \vdots & \vdots & \vdots & \vdots \\
                    1 & h_{N+1} & \cdots & h_{N+1}^N 
                  \end{bmatrix}$ be Vandermonde matrix. Let \begin{equation}\label{eqn:estimation for residual simplified}
\begin{bmatrix}
                                                                       a^e_0 \\
                                                                       a^e_1 \\
                                                                       \vdots \\
                                                                       a^e_N 
                                                                     \end{bmatrix}=W_N^{-1}\begin{bmatrix}
                                                                                             f(x_0) \\
                                                                                             f(x_1) \\
                                                                                             \vdots \\
                                                                                             f(x_N) 
                                                                                           \end{bmatrix}.
\end{equation}
Let  \begin{equation}\label{eqn:matrix representation}
                                                 f_N^e(x)=(1,x-x_0,\cdots,(x-x_0)^N)\begin{bmatrix}
                                                                       a^e_0 \\
                                                                       a^e_1 \\
                                                                       \vdots \\
                                                                       a^e_N 
                                                                     \end{bmatrix}.
                                               \end{equation}
if $\vert x-x_0\vert \le h$, 
Then we have  \begin{equation}\label{eqn:estimate for function representation}
                  \vert f(x)-f_N^e(x) \vert \le  M h^{N+1} ((\frac{ K^{N+1}(K+1)^N }{N!})+ \frac{1}{(N+1)!}).
              \end{equation}

\begin{proof} According Theorem~\ref{thm:Errors of the differential operators}, It yields 
\begin{equation}
  \vert a^e_i -a_i\vert \le \frac{M C_N^{i}K^{2N+1-i} h^{N+1-i}}{N!},
 \end{equation} 
 
 For $\vert x-x_0\vert \le h$, it yields
 \begin{eqnarray}
                  \vert f(x)-f_N^e(x) \vert &=& \vert (\sum_{i=0}^{N} (a_i-a_i^e)(x-x_0)^i) +\frac{f^{(N+1)}(\zeta)}{(N+1)!}(x-x_0)^{N+1}\vert \\
                 &\le& (\sum_{i=0}^{N} \vert a_i-a_i^e\vert \vert x-x_0\vert^i) +\vert\frac{f^{(N+1)}(\zeta)}{(N+1)!}\vert\vert(x-x_0)^{N+1}\vert \\
                 &\le &  (\sum_{i=0}^{N} \frac{M C_N^{i}K^{2N+1-i} h^{N+1-i}}{N!}h^i)+\vert\frac{M}{(N+1)!}\vert\vert h^{N+1}\vert\\
                 &=&  M h^{N+1} ((\sum_{i=0}^{N} \frac{ C_N^{i}K^{2N+1-i}}{N!})+ \frac{1}{(N+1)!}) \\
                 &=& M h^{N+1} ((\frac{ K^{N+1}(K+1)^N }{N!})+ \frac{1}{(N+1)!}).
              \end{eqnarray}
\end{proof}

\end{appendices}

\end{document}